\documentclass{article}

\usepackage{graphicx}
\usepackage{amsmath} 
\usepackage{wrapfig}
\usepackage{float}
\usepackage{tabularx} 
\usepackage{booktabs}
\usepackage{wrapfig}   
\usepackage{amssymb} 

 \usepackage[sglblindworkshop, final]{neurips_2025}

\usepackage[utf8]{inputenc} 
\usepackage[T1]{fontenc}    
\usepackage{hyperref}       
\usepackage{url}            
\usepackage{booktabs}       
\usepackage{amsfonts}       
\usepackage{nicefrac}       
\usepackage{microtype}      
\usepackage{xcolor}         

\title{Avi: A 3D Vision-Language Action Model Architecture \\ generating Action from Volumetric Inference}

\author{%
  Harris Song \\
   University of California, Los Angeles \\
  \And
  Long Le \thanks{Correspondence to: vlongle@seas.upenn.edu} \\
  University of pennsylvania \\
}

\begin{document}

\maketitle

\begin{abstract}
We propose \textbf{Avi}, a novel \textbf{3D Vision-Language-Action (VLA) architecture} that reframes robotic action generation as a problem of 3D perception and spatial reasoning, rather than low-level policy learning. While existing VLA models primarily operate on 2D visual inputs and are trained end-to-end on task-specific action policies, Avi leverages 3D point clouds and language-grounded scene understanding to compute actions through classical geometric transformations. Most notably, \textbf{Avi \textit{does not} train on previous action tokens,} rather, we build upon a 3D Multi-modal Large Language Model (MLLM) to generate the next point cloud and explicitly calculate the actions through classical transformations. This approach enables generalizable behaviors that are robust to occlusions, camera pose variations, and changes in viewpoint. By treating the robotic decision-making process as a structured reasoning task over 3D representations, Avi bridges the gap between high-level language instructions and low-level actuation without requiring opaque policy learning. Our preliminary results highlight the potential of 3D vision-language reasoning as a foundation for scalable, robust robotic systems. Check it out at 
\href{https://avi-3drobot.github.io}{avi-3drobot.github.io}.
\end{abstract}

\section{Introduction}
\begin{figure*}[ht]
    \centering
    \includegraphics[width=\textwidth]{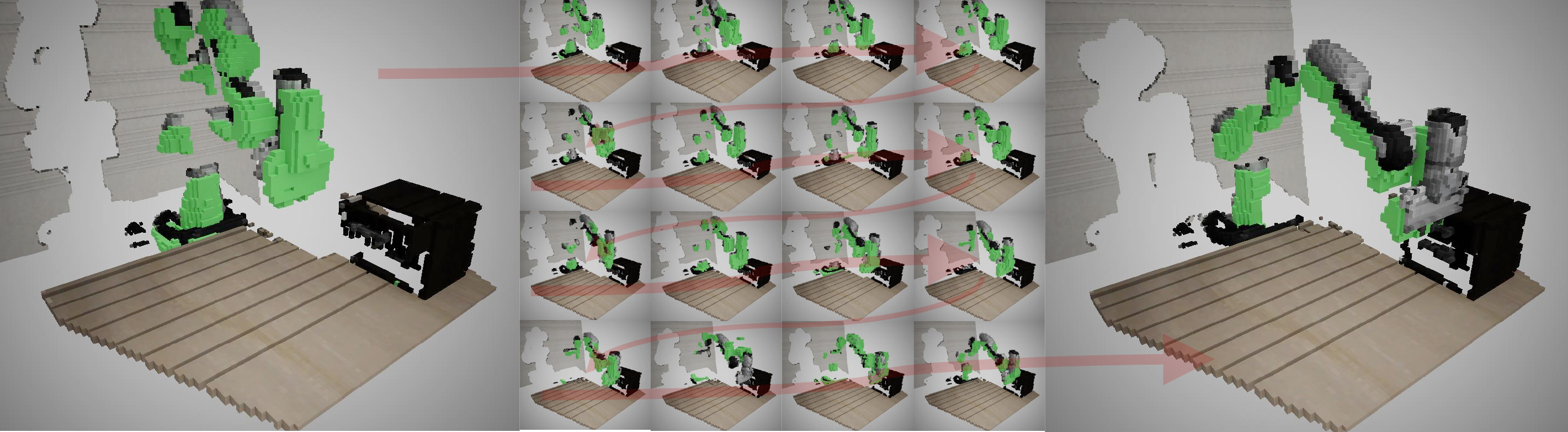}
    \caption{The left image represents the starting position of the scene. The green voxels represent the predicted next time stamp. The right image represents the end time stamp, and the series of images in between indicate the rollout.}
    \label{fig:flow}
\end{figure*}

Vision-Language-Action (VLA) models have recently gained significant attention in the robotics and machine learning communities \cite{black2410pi0, team2025gemini}.  
While these models have demonstrated impressive capabilities in connecting high-level natural language instructions with actionable robot policies, the vast majority of current VLAs operate solely on 2D image inputs \cite{chi2023diffusion, li2025unifiedvideoaction, black2410pi0}. This reliance on 2D perception imposes fundamental limitations: reasoning about depth, object geometry, and fine-grained spatial relations becomes indirect and often error-prone.  

There are other novel Vision-Language Action models that infer based on other sources, such as implicitly 3D Point Clouds \cite{yang2025fp33dfoundationpolicy, cheng2025fvp}, but they treat the robotics problem as an end-to-end action generation problem and implicitly assume that the model is generating action tokens in a \textit{scene and robot-specific} setup, decreasing the reproducibility of these results. 

In this work, we propose to move beyond both the 2D perception and action-based paradigm by training a VLA model that \emph{natively} operates on 3D representations, specifically point clouds.  
Our approach is built upon ShapeLLM-Omni, a 3D Multi-Modal Language Model (3D MMLM), which we finetune to condition on both natural language commands and a 3D point cloud of the scene. \cite{ye2025shapellm} Rather than directly outputting low-level joint actions, the model predicts a \emph{delta point cloud} that represents the desired post-condition of the manipulated object(s). Robot joint actions are then derived through traditional inverse kinematics, aligning the end-effector to the predicted ``after'' state of the point cloud.  

In summary, we present two main contributions:
\begin{enumerate}
    \item \textbf{AVI (Action from Volumetric Inference)}: a novel architecture that integrates a 3D Multi-Modal Language Model to infer actions through volumetric reasoning, rather than directly generating action tokens. \textbf{More importantly, our architecture doesn't require training on previous action tokens, but rather, only previous depth maps}. This approach shifts the focus from language-to-action to language-to-geometry, enabling richer spatial grounding.  
    \item \textbf{Location Quantization for 3D MLLMs}: a general technique for discretizing spatial information that allows pretrained 3D MLLMs to generalize at the \emph{object level} rather than at the scene level. Current state-of-the-art 3D MLLMs, specifically ShapeLLM-Omni \cite{ye2025shapellm}, are built on training assets with online 3D Models rather than entire scene generation. Developing a simple location quantization technique helps us overcome this technical barrier while demonstrating the effectiveness of our architecture.
\end{enumerate}

\section{Related Work}

Our work lies at the intersection of two separate trends. The first one is in robotics, where the increase in large robotics datasets such as Droid, Open X Embodiment has led to the development of many novel architectures \cite{khazatsky2024droid, open_x_embodiment_rt_x_2023}. 

\begin{figure*}[t]
    \centering
    \includegraphics[width=\textwidth]{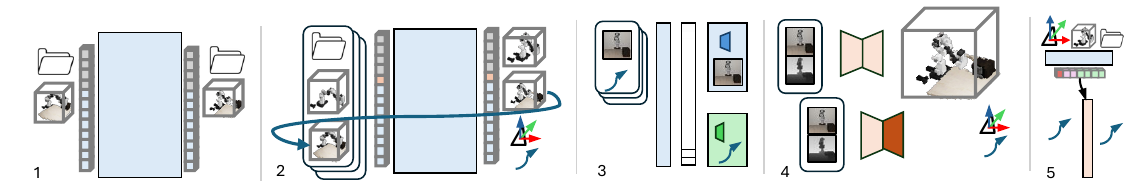}
    \caption{Comparison against related work. (1) describes Shape LLM Omni. (2) describes our work, Avi. (3) describes the Unified Video Action Model. (4) describes Robot 4D Generation. (5) describes 3D Foundation Policy, FP3.}
    \label{fig:comparison}
\end{figure*}
The second trend is within computer vision, where there is now a strong emphasis towards understanding the 3D Space. 3D Reconstruction techniques like NeRF and Gaussian Splatting are capable of learning camera parameters for 3D spatial representation. In addition, image encoders are capable of deriving semantic information, and nowadays, are capable of deriving enough semantic information for reconstruction in the 3D Space.
\subsection{Understanding the Robotics Policy Trend}
There are two additional trends in the robotics space to analyze. The first trend is that most robotics architectures are based on training action tokens. The second robotics-based trend involves injecting signals into visual components, such as through an image encoder that can extract semantic information.
\paragraph{Action Generation}
We usually dictate these problems as a \(A_{t-n}...A_{t-1} \rightarrow A_t\) \cite{zhao2023learning, torabi2018behavioral}. These prior Vision-Language-Action (VLA) methods directly predict robot-specific action tokens, trained to mimic collected demonstrations.  
For example, PerAct \cite{shridhar2022peract}, CLIPort \cite{shridhar2021cliport}, and Transporter Networks \cite{zeng2021transporter} couple visual perception with learned policies tailored to specific robot morphologies.  
Similarly, Unified Video Action \cite{black2410pi0} and related works integrate video-based action representations, but remain dependent on large-scale, robot-specific data.

\paragraph{2D Image Pretraining}

A significant line of research leverages 2D vision pretraining for robotics.  
We treat this as a \(\{A_{t-n}, I_{t-n}\}... \{A_{t-1}, I_{t-1}\}\rightarrow A_{t}\)
Yen-Chen et al. \cite{yen2020learning} demonstrated transfer learning in manipulation policies by reusing COCO-pretrained backbones.  
Radosavovic et al. \cite{radosavovic2023real} applied masked autoencoding to pretrain ViTs on large-scale 2D data before transferring to robotics.  
More recent work has focused on scaling such paradigms: Lift3D \cite{jia2024lift3d} extends 2D representations into implicit depth and point-cloud embeddings, while 3D-MVP \cite{qian20243d} and FVP \cite{cheng2025fvp} extend multiview pretraining to robotic manipulation and 4D video, respectively.  
World-consistent diffusion models \cite{zhang2025world} also highlight the utility of 2D/temporal video modeling as a precursor to structured 3D reasoning.  
While effective, these approaches remain fundamentally limited by their reliance on 2D visual input for geometry.  

More notably, there are a small subset of papers, such as the Unified Video Action Model, that generate \(\{A_t,  I_t\}\) \cite{li2025unifiedvideoaction}. Their use of \textit{image generation} through a diffusion policy is relatively novel, although their policy still depends on a shared latent space with an action diffusion. 

\subsection{Understanding the Trend of 3D Computer Vision}
There are many papers in the 3D Computer Vision space. We discuss a few from the conversion of 2D to 3D space, and then discuss new papers discussing 3D Large Language Models that first use arbitrary 3D environments, such as game assets for their dataset, then shift to using the real-world 3D world models as 3D spaces become more wide-spread.

\paragraph{Generating 3D Environments} InstantSplat introduces a method to generate implicit 3D representations of a scene without explicit camera parameters, and is one of the state-of-the-art Gaussian Splatting papers through it's speed \cite{fan2024instantsplat}. Spatt3r is a zero-shot gaussian splat from uncalibrated image pairs, generating novel views with one or two images \cite{smart2024splatt3r}. 

\paragraph{Understanding the 3D World} Recon++ \cite{qi2023recon} pioneered contrastive representation learning for point clouds.  
ShapeLLM \cite{qi2024shapellm} builds on Recon++ with ChatGPT-4V generated prompts and LLaMA backbones, surpassing PointLLM \cite{xu2024pointllm}.  
JEPA \cite{pointjepa2025} introduces predictive joint embedding architectures, while SUGAR \cite{chen2024sugar} pretrains a transformer encoder from scratch on a massive dataset of 752.2K single objects and 110.7K multi-object scenes.  
Other works explore integrating LMMs with 3D input, such as LLaVA-3D \cite{zhu2024llava}, VoxPoser \cite{huang2023voxposer}, and PointVLA \cite{li2025pointvla}, which directly inject 3D priors into vision-language models.  
These methods highlight the growing consensus that 3D pretraining provides stronger grounding for manipulation than 2D alone.  

\subsection{Intersecting Robotics and 3D Computer Vision}
At the intersection of 3D perception and policy learning, FP3 \cite{yang2025fp33dfoundationpolicy} and DP3 focus on point-cloud-conditioned diffusion policies, pretrained on large robot datasets such as Droid.  
3D-VLA \cite{zhen20243dvla} and SpatialVLA \cite{qu2025spatialvla} extend this direction by predicting volumetric or depth-infused representations for language-conditioned action.  
Recent embodied generalist agents \cite{huang2024embodied} and Gemini Robotics \cite{team2025gemini} scale VLA models across tasks, but remain heavily compute- and data-intensive.  
Meanwhile, industrial efforts such as Google Robotics and Nvidia GR00T are developing proprietary foundational VLA systems at scale.  
These approaches showcase the trend toward 3D-aware VLA, but most remain tied to \emph{action token prediction}, which restricts generality.  

\paragraph{Comparison to Related Architectures} Figure~\ref{fig:comparison} situates our proposed framework in the broader landscape of vision-language-action models. 
Panel (1) highlights ShapeLLM-Omni, which serves as the foundational 3D multi-modal model but was originally trained on single-object assets and thus struggles with multi-object robotic environments. \cite{ye2025shapellm} 
Panel (3) illustrates the \textit{Unified Video Action Model}, which leverages video-to-action representations but remains limited by the lack of explicit 3D reasoning. \cite{li2025unifiedvideoaction}
Panel (4) shows \textit{Robot 4D Generation}, a video-based generative approach constrained to temporal 2D/3D fusion without volumetric grounding. 
Panel (5) depicts the \textit{3D Foundation Policy (FP3)}, a diffusion-based method that directly generates robot actions from point clouds, but does not predict geometric outcomes explicitly. 

In contrast, Panel (2) presents \textbf{Avi}, our proposed architecture, which introduces a fundamentally different perspective: 
instead of generating actions directly, Avi predicts \emph{delta 3D point clouds} conditioned on natural language instructions, 
and then derives executable robot trajectories via geometric optimization. 
This design shifts the paradigm from \textit{language-to-action} to \textit{language-to-geometry}, 
yielding interpretable, morphology-agnostic behaviors that are robust across embodiments. 
The novelty of Avi lies in combining a 3D MLLM backbone with our proposed \emph{location quantization} strategy, 
bridging the gap between large-scale vision-language pretraining and precise, spatially grounded robotic manipulation.

\paragraph{Summary} Prior work in the space spans policy learning in robotics, and the emergence of 3D computer vision helpfully intersects the policy learning in robotics.
Our method diverges shifts from the current policy learning trends by reframing VLA as \textbf{language-to-geometry}: predicting 3D volumetric transformations instead of action tokens. By contrast, our approach emphasizes a \textbf{morphology-agnostic policy}: rather than outputting actions, our model predicts transformed 3D point clouds from which robot-specific trajectories can be computed via inverse kinematics.

\begin{table}[t]
\centering
\caption{Comparison of our approach with related methods in robotic policy learning.}
\label{tab:comparison}
\scriptsize 
\begin{tabularx}{\textwidth}{p{2cm}X X p{2cm} p{1cm}}
\toprule
\textbf{Method} & \textbf{Input Mode} & \textbf{Core Mechanism} & \textbf{3D Point Clouds?} & \textbf{Relies on Actions?} \\
\midrule
This Work (Avi) & Point Clouds + Language & 3D MLLM predicting delta point clouds + IK & \textbf{No} & \textbf{No} \\
\midrule
Unified Video-Action \cite{li2025unifiedvideoaction} & Images + \textit{Action Tokens} & Joint video–action latent modeling & No & Yes \\
\midrule
Diffusion Policy 3D \cite{ze2024_3d_diffusion_policy} & 3D point clouds + \textit{Action Tokens} & Diffusion model over actions conditioned on 3D & Uses 3D conditioning, outputs actions & Yes \\
\midrule
Diffusion Policy \cite{chi2023diffusion} & Images + \textit{Action Tokens} & Diffusion model over actions conditioned in 2D & Uses 2D conditioning, outputs actions & Yes \\
\midrule
3D Foundation Policy \cite{yang2025fp33dfoundationpolicy}& Point Clouds + Language + \textit{Action Tokens} & Diffusion transformer policy pre-trained on 3D & No (actions directly) & Yes \\
\bottomrule
\end{tabularx}
\end{table}

\section{Method}
Our architecture is novel because we take in point clouds, then we output both point clouds and a valid transformation for the robotic system. We start with a 3D represented point cloud, then iterate through the Segment Anything encoder to extract relevant \textit{objects} from our scene, which is denoted in subsection \ref{subsec:objectseg}. We then describe our location quantization method, which is denoted in subsection \ref{subsec:location}. We then discuss the 3D MLLM that runs inference in subsection \ref{sec:mmlm} and describe a classical transformation strategy to calculate actions.
\begin{figure*}[t]
    \centering
    \includegraphics[width=\textwidth]{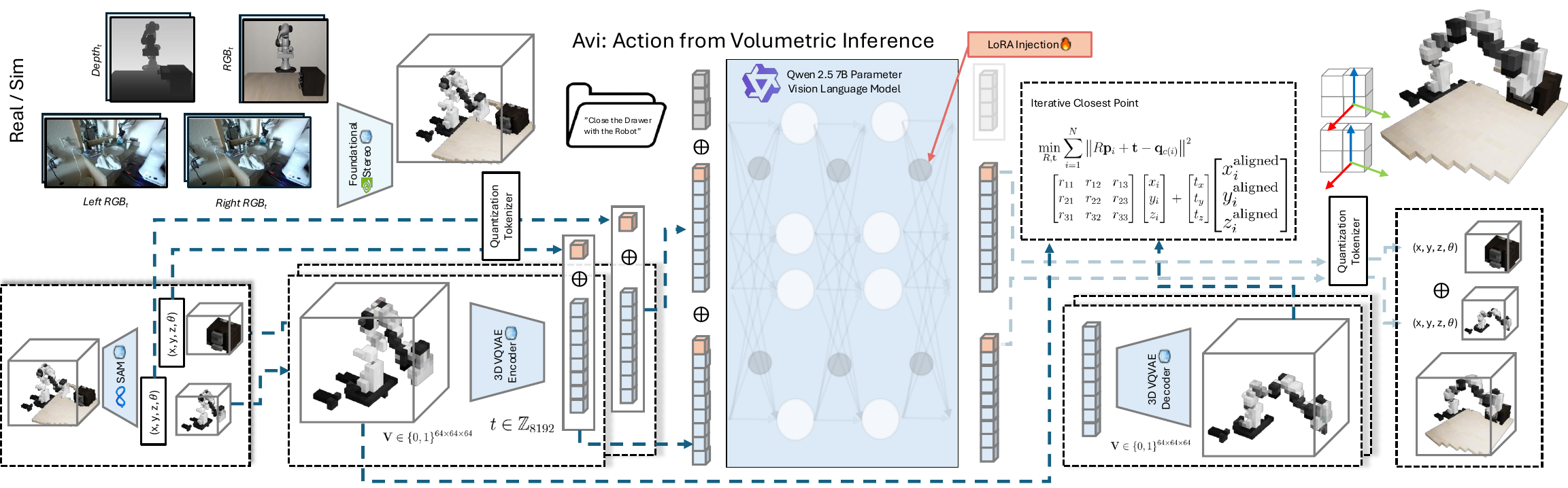}
    \caption{Overview of our Vision-Language Action Model \textbf{Avi}, a volumetric vision-language system for robotic action generation. Avi combines stereo reconstruction, 2D segmentation (via Segment Anything), and a fine-tuned 3D Vision-Language Model (based on Qwen-VL and 3D VQVAE embeddings) to predict goal-conditioned 3D volumes. We further align these volumes using classical geometric optimization (ICP) to produce interpretable, spatially grounded actions.}
    \label{fig:architecture}
\end{figure*}

\subsection{Object Segmentation}
\label{subsec:objectseg}
Formally, we represent each scene as a point cloud 
\[
\mathcal{P} \subset \mathbb{R}^{N \times 3},
\]
where \(N\) denotes the number of sampled points.  
The first step of our framework is to partition $\mathcal{P}$ into a collection of disjoint subsets, each corresponding to an object in the scene:
\[
\mathcal{P} = \bigcup_{k=1}^{K} \mathcal{P}_k, \qquad 
\mathcal{P}_i \cap \mathcal{P}_j = \varnothing \ \ \text{for } i \neq j,
\]
where $\mathcal{P}_k \subset \mathbb{R}^{N_k \times 3}$ denotes the point cloud of the \(k\)-th object, and \(K\) is the number of detected objects.  

To compute this decomposition, we apply a segmentation function 
\[
g_{\text{seg}} : \mathcal{P} \to \{\mathcal{P}_1, \ldots, \mathcal{P}_K\},
\]
which is instantiated via a pretrained image-based segmentation backbone (e.g., the \texttt{Segment Anything Model}) followed by geometric lifting into 3D. Each $\mathcal{P}_k$ thus inherits both semantic labels and fine-grained geometric boundaries from the underlying segmentation.  

Once segmented, each object $\mathcal{P}_k$ is enriched with additional metadata that encodes its spatial properties. Specifically, we associate a tuple of discrete descriptors
\[
\ell_k = (x_k, y_k, z_k, s_k),
\]
where $(x_k,y_k,z_k)$ denotes the quantized centroid location of the object and $s_k$ its quantized scale, as introduced in subsection~\ref{subsec:location}. These descriptors are converted into location tokens and appended to the object-level representation.  

Finally, both geometric and linguistic modalities are embedded into a shared latent space $\mathcal{Z}$.  
Let $\mathcal{T}$ denote the natural language instruction and $\{\mathcal{P}_k, \ell_k\}_{k=1}^K$ the segmented object representations. We define encoders
\[
f_{\text{3D}}(\{\mathcal{P}_k, \ell_k\}_{k=1}^K) \in \mathcal{Z}, 
\qquad f_{\text{text}}(\mathcal{T}) \in \mathcal{Z},
\]
where $f_{\text{3D}}$ integrates both raw geometry and location-token metadata.  
This ensures that the latent space jointly captures semantic intent from language and structured spatial reasoning from 3D geometry, enabling the model to reason over objects rather than raw point distributions.

The token sequence $\mathbf{z}$ is then passed through the VQ-VAE decoder to reconstruct the voxel grid $\hat{\mathcal{V}}$, which is subsequently converted back into a point cloud $\hat{\mathcal{P}} \subset \mathbb{R}^{N \times 3}$.

\subsection{Location Quantization}
\label{subsec:location}

We maintain the initial token embeddings from the previously trained 3D Multi-Modal Large Language Model (3D-MLLM) \emph{ShapeLLM-Omni}, ensuring compatibility with the pretrained architecture.  
To incorporate additional spatial and geometric information, we extend the vocabulary by introducing dedicated \emph{position} and \emph{scale} tokens.  

Specifically, we define three independent position axes: \(
X, Y, Z \in \{1, 2, \dots, 256\},
\) each discretized into \(256\) bins. This introduces a total of 768 new tokens corresponding to positional context. In addition, we discretize the object scale into \(
S \in \{1, 2, \dots, 128\},\) yielding \(128\) scale tokens.   Thus, the overall vocabulary extension equates \(896\) additional tokens. 

Finally, the extended embedding matrix becomes \(
E' \in \mathbb{R}^{(|\mathcal{V}_{0}| + 896) \times d}, \)
where \(|\mathcal{V}_{0}|\) denotes the size of the original vocabulary from ShapeLLM-Omni and \(d\) is the embedding dimension. The new embeddings corresponding to the \(896\) tokens are initialized (e.g., randomly or via scaled normal initialization), while the pretrained embeddings for the original vocabulary are preserved to retain the knowledge of the base model.

\begin{wrapfigure}{r}{0.6\textwidth} 
  \centering
  \includegraphics[width=0.58\textwidth]{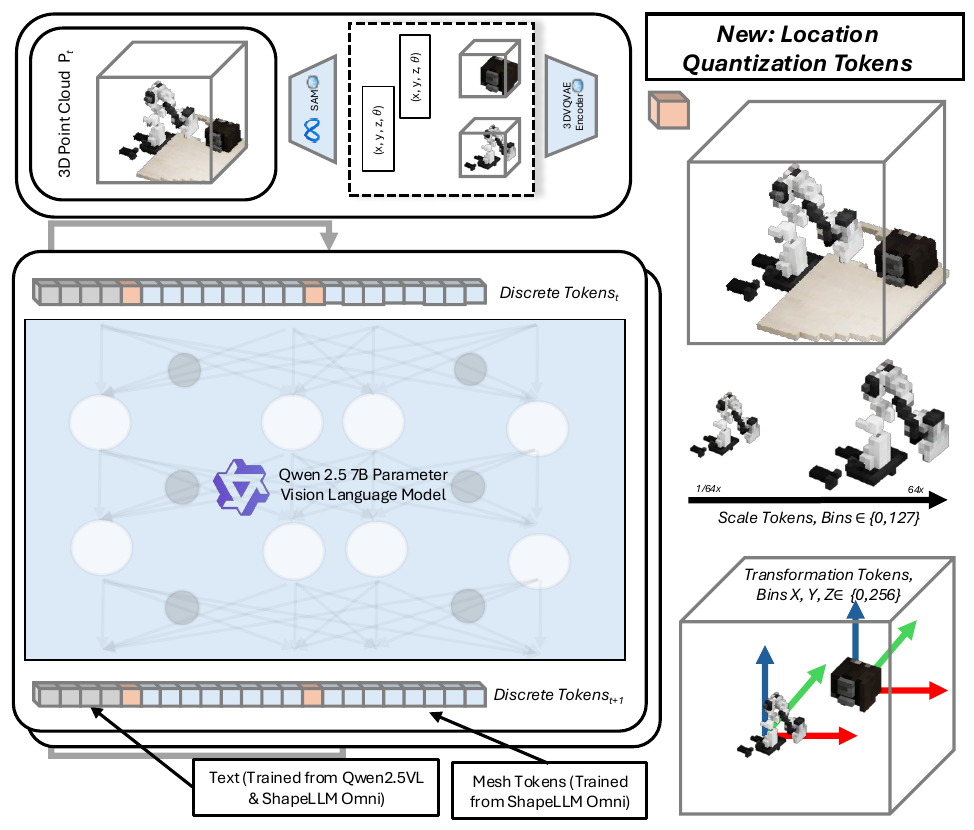}
  \caption{The 3D MLLM training stage with the new extended vocabulary is detailed, along with the discrete tokens represented between \textit{t} and \textit{t+1}. The new Location Quanization tokens are detailed, including the Transformation and Scale Tokens.}
  \label{fig:training}
\end{wrapfigure}

Figure~\ref{fig:training} illustrates the \emph{Location Quantization} method that is incorporated into both the training and inference stages of our framework. For every segmented object, we attach a set of quantized location tokens that encode its spatial context within the 3D environment. Our qualitative ablation studies in Section ~\ref{sec:ablation} discusses the necessity.
These tokens serve as an additional input modality, enabling the model to reason not only about the object’s semantic identity but also about its discretized position and scale relative to the overall scene.  
This mechanism ensures that spatial information is consistently represented throughout the training and evaluation process, thereby enhancing the model’s capacity for grounding and generalization in downstream tasks.

\subsection{Multi-Modal Large Language Model}
\label{sec:mmlm}

We freeze the pretrained VQ-VAE encoder, where VQ-VAE stands for \textit{Vector Quantized Variational Autoencoder}. This encoder maps a voxelized 3D shape $\mathcal{V} \in \mathbb{R}^{64 \times 64 \times 64}$ into a discrete latent representation consisting of 8192 tokens: \(
\text{Encoder}_{\text{VQ-VAE}}(\mathcal{V}) \rightarrow \mathbf{z} = [z_1, z_2, \dots, z_{8192}], \quad z_i \in \mathcal{C} \) where $\mathcal{C}$ is a learned codebook of latent embeddings.

At the core of our framework lies a \emph{Multi-Modal Large Language Model} (MMLM), which extends
traditional autoregressive transformers to jointly reason over 3D geometry and natural language.
Formally, let the model parameters be denoted by $\Theta$, and define a sequence of discrete tokens
$\mathbf{z} = (z_1, z_2, \dots, z_L)$, where each token $z_i$ may originate from one of three modalities:
\[
z_i \in \mathcal{V}_{\text{text}} \cup \mathcal{V}_{\text{3D}} \cup \mathcal{V}_{\text{loc}},
\]
with $\mathcal{V}_{\text{text}}$ the text vocabulary (Qwen-2.5), $\mathcal{V}_{\text{3D}}$ the pretrained
ShapeLLM-Omni codebook, and $\mathcal{V}_{\text{loc}}$ the extended set of position and scale tokens
introduced in Section~\ref{subsec:location}.  

The joint distribution over multimodal tokens is factorized autoregressively:
\[
p_\Theta(\mathbf{z}) = \prod_{i=1}^{L} p_\Theta(z_i \mid z_{<i}),
\]
allowing the model to condition future geometric predictions not only on past geometry tokens,
but also on natural language instructions and quantized spatial context. This design integrates
semantic and spatial reasoning into a unified sequence-modeling framework.  

Concretely, given an instruction $\mathcal{T}$ and scene point cloud $\mathcal{P}$, the encoders
$f_{\text{text}}$ and $f_{\text{3D}}$ map the modalities into a shared embedding space
$\mathcal{Z}$:
\[
h_{\text{text}} = f_{\text{text}}(\mathcal{T}), \qquad
h_{\text{3D}} = f_{\text{3D}}(\mathcal{P}),
\]
which are concatenated with location tokens $h_{\text{loc}}$ to form the input sequence
$\mathbf{h}_0 = [h_{\text{text}}, h_{\text{3D}}, h_{\text{loc}}]$.
The transformer layers then compute contextualized hidden states
$\mathbf{h}_1, \mathbf{h}_2, \dots, \mathbf{h}_L$, which are projected into logits over the
extended vocabulary. This formulation generalizes large language models into a multi-modal setting where both linguistic and geometric reasoning are expressed in the same discrete token space. Importantly, it enables
\emph{language-to-geometry generation}, in which the model autoregressively predicts the next 3D
volumetric state $\hat{\mathcal{P}}_{t+1}$ conditioned on the current state $\mathcal{P}_t$ and prompt
$\mathcal{T}$, thereby bridging semantic instructions with spatial manipulation outcomes.

\subsection{Transformation Calculation}
\label{subsec:transformation}
Given a prompt and current scene point cloud \(\mathcal{P}_t\), we generate a next point cloud prediction \(\hat{\mathcal{P}}_{t+1}\) such that: \(
\hat{\mathcal{P}}_{t+1} \approx \mathcal{P}_t + \Delta \mathcal{P} \) where \(\Delta \mathcal{P}\) represents the learned spatial change conditioned on the prompt. We then compute the Iterative Closest Point (ICP) transformation, defined as the rigid transformation  $(R, t)$ that minimizes the alignment error:

\[
\min_{\mathbf{R}, \mathbf{t}} \sum_{i=1}^{N} \left\| \mathbf{R} \mathbf{x}_i + \mathbf{t} - \mathbf{y}_i \right\|^2
\]

where $\{\mathbf{x}_i\}$ are points from the source point cloud and $\{\mathbf{y}_i\}$ are their closest points in the target point cloud, $\mathbf{R} \in SO(3)$ is a rotation matrix, and $\mathbf{t} \in \mathbb{R}^3$ is a translation vector. The resulting transformation is then applied to the robot's end effector position $(X, Y, Z)$, and the updated pose is executed.

\section{Setup and Experimentation}

We fine-tune the foundational model on robotics training data using a single NVIDIA A6000 GPU with 48GB of memory, which is sufficient to accommodate the pretrained ShapeLLM-Omni backbone.  For training, we utilize the LIBERO Dataset \cite{liu2023libero}, which provides diverse task demonstrations within the Robosuite environment.  
Each demonstration contains synchronized RGB-D observations and robot proprioceptive states for a Franka Panda manipulator.

\begin{wrapfigure}{r}{0.4\textwidth}  
    \centering
    \includegraphics[width=0.38\textwidth]{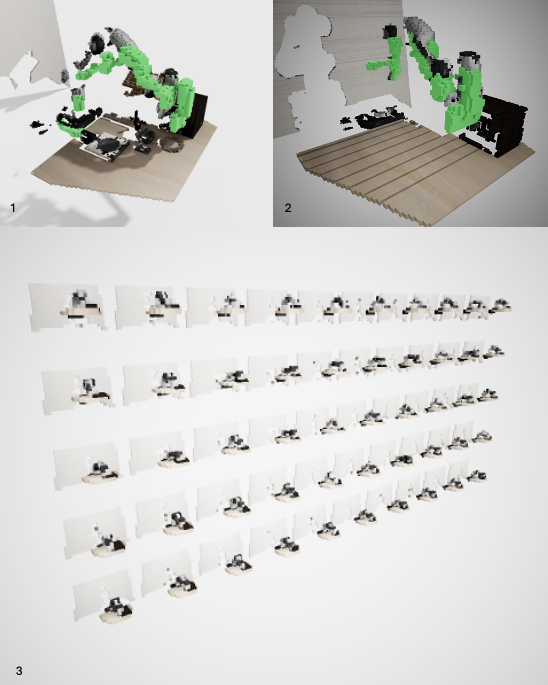}
    \caption{The demonstration expert rollouts provided by LIBERO Goal.}
    \label{fig:training_demonstration}
\end{wrapfigure}

In our experiments, we select \(\,50\) demonstrations corresponding to the drawer-closing task.  
Figure~\ref{fig:flow} illustrates eighteen sample inference rollouts for this task, where the robot is required to align its end-effector and successfully close the drawer.  
 
For segmentation, we employ the Segment Anything Model (SAM) to isolate individual objects from raw visual inputs \cite{kirillov2023segment}. To maintain stability and avoid overfitting, the SAM encoder weights are frozen throughout training, providing consistent object-level features while the rest of the model adapts to robotics-specific tasks.  

To regularize training in this limited-data regime, we apply dropout with a probability of \(p = 0.05\).  
Fine-tuning is performed using Low-Rank Adaptation (LoRA) to enable efficient parameter updates without full model retraining.  
Specifically, we unfreeze the last \(K\) layers of the attention mechanism, inserting LoRA adaptation matrices into the Query (\(Q\)), Key (\(K\)), and Value (\(V\)) projection layers.  
This design allows the model to adapt effectively to manipulation while retaining the broad multimodal reasoning capabilities of the pretrained 3D MLLM.  
To enable efficient fine-tuning, we adopt Low-Rank Adaptation (LoRA) across the attention layers. 
We experiment with rank values 
$r \in \{4, 8, 16, 32, 64\}$ 
to evaluate the sensitivity of the model to the adaptation capacity.  Following common practice, we set the scaling factor proportional to the rank, with 
$\text{LoRA}_{\alpha} = 2r$, 
ensuring that the effective update magnitude grows consistently with model capacity. 
This design allows us to systematically explore the trade-off between computational cost (smaller $r$) 
and representational flexibility (larger $r$), providing insights into how LoRA rank influences 
fine-grained 3D manipulation performance.

\section{Results}

\begin{table}[t]
\centering
\caption{Location quantization ablation. Token-space resolution is $B^3$; world-space perceived scene size is $R_{\text{eff}}^3$ with $R_{\text{eff}}=\max(B, V/s)$, using $V{=}64$ and (here) $s{=}1$.}
\label{tab:lq_ablation_wide}
\setlength{\tabcolsep}{8pt}
\begin{tabular}{lcccc}
\toprule
 & \textbf{No LQ} & \textbf{LQ (64)} & \textbf{LQ (128)} & \textbf{LQ (256)} \\
\midrule
Bins $B$ & -- & 64 & 128 & 256 \\
Token-space resolution $B^3$ & -- & $64^3$ & $128^3$ & $256^3$ \\
VQ-VAE grid $V^3$ & $64^3$ & $64^3$ & $64^3$ & $64^3$ \\
\textbf{Perceived scene size} $R_{\text{eff}}^3$ (world) & $64^3$ & $64^3$ & $128^3$ & $256^3$ \\
Gripper LLM Gen. & No & No & Yes & Yes \\
Qualitative obs. & See Fig.~\ref{fig:detail} & N/A & N/A & See Fig.~\ref{fig:detail} \\
\bottomrule
\end{tabular}

\vspace{0.35em}
\small
\emph{Note.} If objects occupy a fraction $s{<}1$ of the workspace along an axis, then $R_{\text{eff}}=\max(B, V/s)$.
For example, with $s{=}0.5$ and $V{=}64$, $V/s{=}128$ so even \textbf{No LQ} attains $128^3$ world-space detail \emph{within the object’s region}. Increasing $B$ refines placement granularity across the workspace.
\end{table}

\begin{wraptable}{r}{0.42\textwidth} 
\vspace{-1.2em}
\centering
\caption{Mean success rate per policy and scene (aggregate over twenty rollouts).}
\label{tab:kitchen_summary}
\small 
\setlength{\tabcolsep}{4pt} 
\begin{tabular}{p{2.2cm}cc}
\toprule
\textbf{Policy} & \textbf{Scene 5} & \textbf{Scene 10} \\
\midrule
ResNet--RNN \cite{liu2023libero} & 0.05 $\pm$ 0.07 & 0.45 $\pm$ 0.11 \\
ResNet--T   \cite{liu2023libero} & 0.80 $\pm$ 0.09 & 0.45 $\pm$ 0.11 \\
ViT--T      \cite{liu2023libero} & 0.90 $\pm$ 0.07 & 0.60 $\pm$ 0.10 \\
Diffusion Policy \cite{chi2023diffusion} & 0.85 $\pm$ 0.08 & 0.70 $\pm$ 0.10 \\
Avi (ours)  & 0.90 $\pm$ 0.07 & 0.90 $\pm$ 0.07 \\
\bottomrule
\end{tabular}

\end{wraptable}

We present preliminary results of our proposed architecture through Figure \ref{fig:flow}, Figure \ref{fig:comparison}, Figure \ref{fig:training_demonstration} and Table \ref{tab:kitchen_summary}. Figure~\ref{fig:flow} illustrates the rollout of the drawer-closing task across eighteen inference steps. The leftmost panel depicts the initial state of the scene, while the rightmost panel shows the final state after execution. 
Intermediate frames visualize the predicted voxelized ``delta'' states (shown in green), which are progressively aligned with the ground-truth trajectory. 
These results demonstrate that Avi is able to generate semantically consistent and physically realizable action trajectories conditioned on natural language instructions.  

Figure \ref{fig:training_demonstration} demonstrates the two demonstrations, where Panel 1 indicates \textit{Scene 10} and Panel 2 indicates \textit{Scene 5}. Neither the Diffusion Policy nor our policy was changed on the domain shift for \textit{Scene 10}. Table \ref{tab:kitchen_summary} represents the success rate of the task rolled out twenty times. Notably, the task \textit{Scene 10} introduces various tabletop items that were not present in \textit{Scene 5}, and highlights the robust nature of the architecture in more complex environments where the number of objects change, but the task does not. It is important that more rollouts are planned, and more rollouts per scene will be required to decrease the variance.
\begin{figure*}[t]
    \centering
    \includegraphics[width=\textwidth]{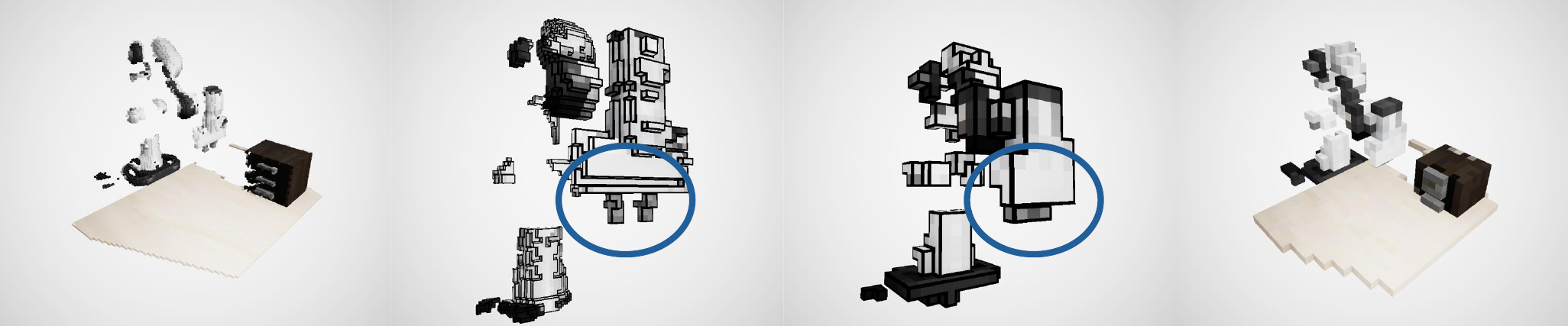}
    \caption{The \textbf{Left Images} contain the architecture with location quantization, including both the world scene and the robot. The \textbf{Right Images} include the architecture without location quantization.}
    \label{fig:detail}
\end{figure*}
\paragraph{Failure Conditions}
Figure ~\ref{fig:flow} contains inferences which highlight when there are failure conditions. For example, the inference sometimes generates points clouds that are out of step and therefore the classical transformation calculator detailed in subsection \ref{subsec:transformation} will calculate incorrect transformations for the gripper. 
\subsection{Ablation Studies}
\label{sec:ablation}
Figure~\ref{fig:detail} presents a side-by-side comparison of the location quantization; the left panel illustrates our full architecture with location quantization enabled, while the right panel shows the same model architecture without this component.  
\begin{wraptable}{r}{0.55\textwidth}
\vspace{-1.2em}
\centering
\scriptsize
\caption{Finetuning results across number of observations and rank $r$. $\alpha$ is calculated as $2r$}
\label{tab:finetuning}
\vspace{2mm}
\begin{tabular}{c|ccccc}
\toprule
\# Obs. & $r=4$ & $r=8$ & $r=16$ & $r=32$ & $r=64$ \\
\midrule
1   & 0.0031  & 0.00001  & 0.00001 & ---- & ---- \\
10  & 0.0091  & 0.00012 & ---- & ---- & ---- \\
25  & 0.00109 & 0.00019 & 0.00001 & 0.00001 & ---- \\
50  & 0.03271 & 0.00241 & 0.00003 & 0.00001 & 0.00001 \\
100 & 0.03418 & 0.00391 & 0.000071& 0.00004 & 0.00001 \\
200 & 0.03912 & 0.00392 & 0.00312 & 0.000217& 0.00002 \\
\bottomrule
\end{tabular}
\vspace{-0.8em}
\end{wraptable}

With location quantization, the model leverages discretized spatial embeddings to ground geometric reasoning, resulting in more reliable alignment of the gripper with the target object and more consistent execution of precision tasks. Precision is increased, as shown from Table \ref{tab:lq_ablation_wide}, highlighting the number of voxels generated per scene is substantially increased despite a small increase to the vocabulary.

In addition, generating our results required a large number rank $r$, The training loss in generating our current model is highlighted. Note that each example is trained to a hundred epochs, following the experimental requirements and hardware detailed above. Training for a hundred epochs took approximately two hours for fifty observations, and the rank size $r$ did not noticeably increase the training time. Note that although the loss is relatively small, we require a higher precision due to the VQVAE tokenization strategy in the architecture. We discuss different loss strategies and future work in Section \ref{sec:conclusion}, such as unfreezing the VQVAE and gradient flow.

\section{Conclusion}
\label{sec:conclusion}

In this work, we introduced \textbf{Avi}, a novel 3D Vision-Language-Action (VLA) architecture that reframes robotic control as a problem of volumetric reasoning rather than low-level policy generation.  
By leveraging ShapeLLM-Omni as a 3D Multi-Modal Language Model and extending it with location quantization, we enable the model to interpret natural language instructions and predict goal-conditioned 3D representations of the environment.  
These predicted volumes are then aligned through geometric optimization, yielding interpretable and morphology-agnostic actions.  
\paragraph{Limitations} There are a few limitations for our work. First, we focus on explicitly generating the next point cloud, but this fails to consider the long-horizon high-level planning that is essential to current papers in the space. Second, our use of a base auto-regressive transformer model is also of concern, and was used as a representation of a 3D MLLM capable of \textit{semantic understanding and generation} in one model. There is currently promising work by SpatialVerse that are working on stronger 3D MLLMs capable of entire scene generation \cite{SpatialGen2025}. Current 3D MLLMs lack access to large-scale training data, especially when compared to state-of-the-art video generation methods. As illustrated in Figure~\ref{fig:flow}, this constraint can lead the baseline model to occasionally generate incorrect frames. At present, we optimize using cross-entropy loss, but additional loss functions merit exploration. For example, unfreezing the 3D VQVAE could provide richer supervision, though at the cost of significantly increased computational requirements.

\paragraph{Future Work} Future work could include developing a novel diffusion-policy that removes the quantization between time frames, albeit in the 3D Space. There is current work that actually does something very similar, albeit they are sidestepping the 3D video generation problem through combining existing diffusion-based autoregressive video generators \cite{liu2025geometryaware4d}. Future work could focus on integrating stronger 3D MLLMs, as our proposed location quantization strategy was developed to address inherent limitations in the current backbone. Beyond architectural improvements, incorporating 3D diffusion-based generative models presents a promising direction. Diffusion models may be particularly well-suited for robotics, since their iterative refinement process can be interpreted as a form of imitation learning, in contrast to the purely autoregressive behavior of Transformer-based approaches. Recent work, such as Masked Autoregressive (MAR) video generation, demonstrates the effectiveness of diffusion-style losses in place of cross-entropy, suggesting that similar techniques could enhance spatial reasoning and long-horizon action generation in 3D VLA systems.

\bibliographystyle{plainnat}  
\bibliography{neurips}           

\newpage
\section*{NeurIPS Paper Checklist}

The checklist is designed to encourage best practices for responsible machine learning research, addressing issues of reproducibility, transparency, research ethics, and societal impact. Do not remove the checklist: {\bf The papers not including the checklist will be desk rejected.} The checklist should follow the references and follow the (optional) supplemental material.  The checklist does NOT count towards the page
limit. 

Please read the checklist guidelines carefully for information on how to answer these questions. For each question in the checklist:
\begin{itemize}
    \item You should answer \answerYes{}, \answerNo{}, or \answerNA{}.
    \item \answerNA{} means either that the question is Not Applicable for that particular paper or the relevant information is Not Available.
    \item Please provide a short (1–2 sentence) justification right after your answer (even for NA). 
\end{itemize}

{\bf The checklist answers are an integral part of your paper submission.} They are visible to the reviewers, area chairs, senior area chairs, and ethics reviewers. You will be asked to also include it (after eventual revisions) with the final version of your paper, and its final version will be published with the paper.

The reviewers of your paper will be asked to use the checklist as one of the factors in their evaluation. While "\answerYes{}" is generally preferable to "\answerNo{}", it is perfectly acceptable to answer "\answerNo{}" provided a proper justification is given (e.g., "error bars are not reported because it would be too computationally expensive" or "we were unable to find the license for the dataset we used"). In general, answering "\answerNo{}" or "\answerNA{}" is not grounds for rejection. While the questions are phrased in a binary way, we acknowledge that the true answer is often more nuanced, so please just use your best judgment and write a justification to elaborate. All supporting evidence can appear either in the main paper or the supplemental material, provided in appendix. If you answer \answerYes{} to a question, in the justification please point to the section(s) where related material for the question can be found.

IMPORTANT, please:
\begin{itemize}
    \item {\bf Delete this instruction block, but keep the section heading ``NeurIPS Paper Checklist"},
    \item  {\bf Keep the checklist subsection headings, questions/answers and guidelines below.}
    \item {\bf Do not modify the questions and only use the provided macros for your answers}.
\end{itemize}


\begin{enumerate}

\item {\bf Claims}
    \item[] Question: Do the main claims made in the abstract and introduction accurately reflect the paper's contributions and scope?
    \item[] Answer: \answerYes{}{} 
    \item[] Justification: The main claims made in the abstract and introduction accurately reflects the paper's contribution and scope.
    \item[] Guidelines:
    \begin{itemize}
        \item The answer NA means that the abstract and introduction do not include the claims made in the paper.
        \item The abstract and/or introduction should clearly state the claims made, including the contributions made in the paper and important assumptions and limitations. A No or NA answer to this question will not be perceived well by the reviewers. 
        \item The claims made should match theoretical and experimental results, and reflect how much the results can be expected to generalize to other settings. 
        \item It is fine to include aspirational goals as motivation as long as it is clear that these goals are not attained by the paper. 
    \end{itemize}

\item {\bf Limitations}
    \item[] Question: Does the paper discuss the limitations of the work performed by the authors?
    \item[] Answer: \answerYes{} 
    \item[] Justification: Yes, we include a section that describes the limitations and further work described by the authors.
    \item[] Guidelines:
    \begin{itemize}
        \item The answer NA means that the paper has no limitation while the answer No means that the paper has limitations, but those are not discussed in the paper. 
        \item The authors are encouraged to create a separate "Limitations" section in their paper.
        \item The paper should point out any strong assumptions and how robust the results are to violations of these assumptions (e.g., independence assumptions, noiseless settings, model well-specification, asymptotic approximations only holding locally). The authors should reflect on how these assumptions might be violated in practice and what the implications would be.
        \item The authors should reflect on the scope of the claims made, e.g., if the approach was only tested on a few datasets or with a few runs. In general, empirical results often depend on implicit assumptions, which should be articulated.
        \item The authors should reflect on the factors that influence the performance of the approach. For example, a facial recognition algorithm may perform poorly when image resolution is low or images are taken in low lighting. Or a speech-to-text system might not be used reliably to provide closed captions for online lectures because it fails to handle technical jargon.
        \item The authors should discuss the computational efficiency of the proposed algorithms and how they scale with dataset size.
        \item If applicable, the authors should discuss possible limitations of their approach to address problems of privacy and fairness.
        \item While the authors might fear that complete honesty about limitations might be used by reviewers as grounds for rejection, a worse outcome might be that reviewers discover limitations that aren't acknowledged in the paper. The authors should use their best judgment and recognize that individual actions in favor of transparency play an important role in developing norms that preserve the integrity of the community. Reviewers will be specifically instructed to not penalize honesty concerning limitations.
    \end{itemize}

\item {\bf Theory assumptions and proofs}
    \item[] Question: For each theoretical result, does the paper provide the full set of assumptions and a complete (and correct) proof?
    \item[] Answer: \answerNA{} 
    \item[] Justification: The paper does not include theoretical results.
    \item[] Guidelines:
    \begin{itemize}
        \item The answer NA means that the paper does not include theoretical results. 
        \item All the theorems, formulas, and proofs in the paper should be numbered and cross-referenced.
        \item All assumptions should be clearly stated or referenced in the statement of any theorems.
        \item The proofs can either appear in the main paper or the supplemental material, but if they appear in the supplemental material, the authors are encouraged to provide a short proof sketch to provide intuition. 
        \item Inversely, any informal proof provided in the core of the paper should be complemented by formal proofs provided in appendix or supplemental material.
        \item Theorems and Lemmas that the proof relies upon should be properly referenced. 
    \end{itemize}

    \item {\bf Experimental result reproducibility}
    \item[] Question: Does the paper fully disclose all the information needed to reproduce the main experimental results of the paper to the extent that it affects the main claims and/or conclusions of the paper (regardless of whether the code and data are provided or not)?
    \item[] Answer: \answerYes{} 
    \item[] Justification: Yes, the paper discloses the steps required to reproduce the experimental results of the paper.
    \item[] Guidelines:
    \begin{itemize}
        \item The answer NA means that the paper does not include experiments.
        \item If the paper includes experiments, a No answer to this question will not be perceived well by the reviewers: Making the paper reproducible is important, regardless of whether the code and data are provided or not.
        \item If the contribution is a dataset and/or model, the authors should describe the steps taken to make their results reproducible or verifiable. 
        \item Depending on the contribution, reproducibility can be accomplished in various ways. For example, if the contribution is a novel architecture, describing the architecture fully might suffice, or if the contribution is a specific model and empirical evaluation, it may be necessary to either make it possible for others to replicate the model with the same dataset, or provide access to the model. In general. releasing code and data is often one good way to accomplish this, but reproducibility can also be provided via detailed instructions for how to replicate the results, access to a hosted model (e.g., in the case of a large language model), releasing of a model checkpoint, or other means that are appropriate to the research performed.
        \item While NeurIPS does not require releasing code, the conference does require all submissions to provide some reasonable avenue for reproducibility, which may depend on the nature of the contribution. For example
        \begin{enumerate}
            \item If the contribution is primarily a new algorithm, the paper should make it clear how to reproduce that algorithm.
            \item If the contribution is primarily a new model architecture, the paper should describe the architecture clearly and fully.
            \item If the contribution is a new model (e.g., a large language model), then there should either be a way to access this model for reproducing the results or a way to reproduce the model (e.g., with an open-source dataset or instructions for how to construct the dataset).
            \item We recognize that reproducibility may be tricky in some cases, in which case authors are welcome to describe the particular way they provide for reproducibility. In the case of closed-source models, it may be that access to the model is limited in some way (e.g., to registered users), but it should be possible for other researchers to have some path to reproducing or verifying the results.
        \end{enumerate}
    \end{itemize}

\item {\bf Open access to data and code}
    \item[] Question: Does the paper provide open access to the data and code, with sufficient instructions to faithfully reproduce the main experimental results, as described in supplemental material?
    \item[] Answer: \answerYes{} 
    \item[] Justification: Yes, the paper will provide open access to the code. The data comes from the another work, the LIBERO dataset. 
    \item[] Guidelines:
    \begin{itemize}
        \item The answer NA means that paper does not include experiments requiring code.
        \item Please see the NeurIPS code and data submission guidelines (\url{https://nips.cc/public/guides/CodeSubmissionPolicy}) for more details.
        \item While we encourage the release of code and data, we understand that this might not be possible, so “No” is an acceptable answer. Papers cannot be rejected simply for not including code, unless this is central to the contribution (e.g., for a new open-source benchmark).
        \item The instructions should contain the exact command and environment needed to run to reproduce the results. See the NeurIPS code and data submission guidelines (\url{https://nips.cc/public/guides/CodeSubmissionPolicy}) for more details.
        \item The authors should provide instructions on data access and preparation, including how to access the raw data, preprocessed data, intermediate data, and generated data, etc.
        \item The authors should provide scripts to reproduce all experimental results for the new proposed method and baselines. If only a subset of experiments are reproducible, they should state which ones are omitted from the script and why.
        \item At submission time, to preserve anonymity, the authors should release anonymized versions (if applicable).
        \item Providing as much information as possible in supplemental material (appended to the paper) is recommended, but including URLs to data and code is permitted.
    \end{itemize}

\item {\bf Experimental setting/details}
    \item[] Question: Does the paper specify all the training and test details (e.g., data splits, hyperparameters, how they were chosen, type of optimizer, etc.) necessary to understand the results?
    \item[] Answer: \answerYes{} 
    \item[] Justification: Yes, hyperparameters were specified in the Methods sections.
    \item[] Guidelines:
    \begin{itemize}
        \item The answer NA means that the paper does not include experiments.
        \item The experimental setting should be presented in the core of the paper to a level of detail that is necessary to appreciate the results and make sense of them.
        \item The full details can be provided either with the code, in appendix, or as supplemental material.
    \end{itemize}

\item {\bf Experiment statistical significance}
    \item[] Question: Does the paper report error bars suitably and correctly defined or other appropriate information about the statistical significance of the experiments?
    \item[] Answer: \answerNA{} 
    \item[] Justification: Experimental statistical significance is expected in a final version of the paper.
    \item[] Guidelines:
    \begin{itemize}
        \item The answer NA means that the paper does not include experiments.
        \item The authors should answer "Yes" if the results are accompanied by error bars, confidence intervals, or statistical significance tests, at least for the experiments that support the main claims of the paper.
        \item The factors of variability that the error bars are capturing should be clearly stated (for example, train/test split, initialization, random drawing of some parameter, or overall run with given experimental conditions).
        \item The method for calculating the error bars should be explained (closed form formula, call to a library function, bootstrap, etc.)
        \item The assumptions made should be given (e.g., Normally distributed errors).
        \item It should be clear whether the error bar is the standard deviation or the standard error of the mean.
        \item It is OK to report 1-sigma error bars, but one should state it. The authors should preferably report a 2-sigma error bar than state that they have a 96\% CI, if the hypothesis of Normality of errors is not verified.
        \item For asymmetric distributions, the authors should be careful not to show in tables or figures symmetric error bars that would yield results that are out of range (e.g. negative error rates).
        \item If error bars are reported in tables or plots, The authors should explain in the text how they were calculated and reference the corresponding figures or tables in the text.
    \end{itemize}

\item {\bf Experiments compute resources}
    \item[] Question: For each experiment, does the paper provide sufficient information on the computer resources (type of compute workers, memory, time of execution) needed to reproduce the experiments?
    \item[] Answer: \answerYes{} 
    \item[] Justification: Yes, the paper provides sufficient information on the resources required for training and inference.
    \item[] Guidelines:
    \begin{itemize}
        \item The answer NA means that the paper does not include experiments.
        \item The paper should indicate the type of compute workers CPU or GPU, internal cluster, or cloud provider, including relevant memory and storage.
        \item The paper should provide the amount of compute required for each of the individual experimental runs as well as estimate the total compute. 
        \item The paper should disclose whether the full research project required more compute than the experiments reported in the paper (e.g., preliminary or failed experiments that didn't make it into the paper). 
    \end{itemize}
    
\item {\bf Code of ethics}
    \item[] Question: Does the research conducted in the paper conform, in every respect, with the NeurIPS Code of Ethics \url{https://neurips.cc/public/EthicsGuidelines}?
    \item[] Answer: \answerYes{} 
    \item[] Justification: Yes, the reviewers have conformed with the NeurIPS Code of Ethics. 
    \item[] Guidelines:
    \begin{itemize}
        \item The answer NA means that the authors have not reviewed the NeurIPS Code of Ethics.
        \item If the authors answer No, they should explain the special circumstances that require a deviation from the Code of Ethics.
        \item The authors should make sure to preserve anonymity (e.g., if there is a special consideration due to laws or regulations in their jurisdiction).
    \end{itemize}

\item {\bf Broader impacts}
    \item[] Question: Does the paper discuss both potential positive societal impacts and negative societal impacts of the work performed?
    \item[] Answer: \answerYes{} 
    \item[] Justification: Yes, broader effects of this work are discussed in the conclusion.
    \item[] Guidelines:
    \begin{itemize}
        \item The answer NA means that there is no societal impact of the work performed.
        \item If the authors answer NA or No, they should explain why their work has no societal impact or why the paper does not address societal impact.
        \item Examples of negative societal impacts include potential malicious or unintended uses (e.g., disinformation, generating fake profiles, surveillance), fairness considerations (e.g., deployment of technologies that could make decisions that unfairly impact specific groups), privacy considerations, and security considerations.
        \item The conference expects that many papers will be foundational research and not tied to particular applications, let alone deployments. However, if there is a direct path to any negative applications, the authors should point it out. For example, it is legitimate to point out that an improvement in the quality of generative models could be used to generate deepfakes for disinformation. On the other hand, it is not needed to point out that a generic algorithm for optimizing neural networks could enable people to train models that generate Deepfakes faster.
        \item The authors should consider possible harms that could arise when the technology is being used as intended and functioning correctly, harms that could arise when the technology is being used as intended but gives incorrect results, and harms following from (intentional or unintentional) misuse of the technology.
        \item If there are negative societal impacts, the authors could also discuss possible mitigation strategies (e.g., gated release of models, providing defenses in addition to attacks, mechanisms for monitoring misuse, mechanisms to monitor how a system learns from feedback over time, improving the efficiency and accessibility of ML).
    \end{itemize}
    
\item {\bf Safeguards}
    \item[] Question: Does the paper describe safeguards that have been put in place for responsible release of data or models that have a high risk for misuse (e.g., pretrained language models, image generators, or scraped datasets)?
    \item[] Answer: \answerYes{} 
    \item[] Justification: The paper discusses limitations within the Conclusion, and a limitation includes the LLM generating incorrect content that should be properly accounted for in the real world.
    \item[] Guidelines:
    \begin{itemize}
        \item The answer NA means that the paper poses no such risks.
        \item Released models that have a high risk for misuse or dual-use should be released with necessary safeguards to allow for controlled use of the model, for example by requiring that users adhere to usage guidelines or restrictions to access the model or implementing safety filters. 
        \item Datasets that have been scraped from the Internet could pose safety risks. The authors should describe how they avoided releasing unsafe images.
        \item We recognize that providing effective safeguards is challenging, and many papers do not require this, but we encourage authors to take this into account and make a best faith effort.
    \end{itemize}

\item {\bf Licenses for existing assets}
    \item[] Question: Are the creators or original owners of assets (e.g., code, data, models), used in the paper, properly credited and are the license and terms of use explicitly mentioned and properly respected?
    \item[] Answer: \answerYes{} 
    \item[] Justification: Yes, proper citations and references to the dataset are used.
    \item[] Guidelines:
    \begin{itemize}
        \item The answer NA means that the paper does not use existing assets.
        \item The authors should cite the original paper that produced the code package or dataset.
        \item The authors should state which version of the asset is used and, if possible, include a URL.
        \item The name of the license (e.g., CC-BY 4.0) should be included for each asset.
        \item For scraped data from a particular source (e.g., website), the copyright and terms of service of that source should be provided.
        \item If assets are released, the license, copyright information, and terms of use in the package should be provided. For popular datasets, \url{paperswithcode.com/datasets} has curated licenses for some datasets. Their licensing guide can help determine the license of a dataset.
        \item For existing datasets that are re-packaged, both the original license and the license of the derived asset (if it has changed) should be provided.
        \item If this information is not available online, the authors are encouraged to reach out to the asset's creators.
    \end{itemize}

\item {\bf New assets}
    \item[] Question: Are new assets introduced in the paper well documented and is the documentation provided alongside the assets?
    \item[] Answer: \answerYes{} 
    \item[] Justification: Yes, new assets introduced in the paper will be released.
    \item[] Guidelines:
    \begin{itemize}
        \item The answer NA means that the paper does not release new assets.
        \item Researchers should communicate the details of the dataset/code/model as part of their submissions via structured templates. This includes details about training, license, limitations, etc. 
        \item The paper should discuss whether and how consent was obtained from people whose asset is used.
        \item At submission time, remember to anonymize your assets (if applicable). You can either create an anonymized URL or include an anonymized zip file.
    \end{itemize}

\item {\bf Crowdsourcing and research with human subjects}
    \item[] Question: For crowdsourcing experiments and research with human subjects, does the paper include the full text of instructions given to participants and screenshots, if applicable, as well as details about compensation (if any)? 
    \item[] Answer: \answerNA{} 
    \item[] Justification: The paper does not involve crowdsourcing nor research with human subjects.
    \item[] Guidelines:
    \begin{itemize}
        \item The answer NA means that the paper does not involve crowdsourcing nor research with human subjects.
        \item Including this information in the supplemental material is fine, but if the main contribution of the paper involves human subjects, then as much detail as possible should be included in the main paper. 
        \item According to the NeurIPS Code of Ethics, workers involved in data collection, curation, or other labor should be paid at least the minimum wage in the country of the data collector. 
    \end{itemize}

\item {\bf Institutional review board (IRB) approvals or equivalent for research with human subjects}
    \item[] Question: Does the paper describe potential risks incurred by study participants, whether such risks were disclosed to the subjects, and whether Institutional Review Board (IRB) approvals (or an equivalent approval/review based on the requirements of your country or institution) were obtained?
    \item[] Answer: \answerNA{} 
    \item[] Justification: The paper does not involved crowdsourcing nor research with human subjects.
    \item[] Guidelines:
    \begin{itemize}
        \item The answer NA means that the paper does not involve crowdsourcing nor research with human subjects.
        \item Depending on the country in which research is conducted, IRB approval (or equivalent) may be required for any human subjects research. If you obtained IRB approval, you should clearly state this in the paper. 
        \item We recognize that the procedures for this may vary significantly between institutions and locations, and we expect authors to adhere to the NeurIPS Code of Ethics and the guidelines for their institution. 
        \item For initial submissions, do not include any information that would break anonymity (if applicable), such as the institution conducting the review.
    \end{itemize}

\item {\bf Declaration of LLM usage}
    \item[] Question: Does the paper describe the usage of LLMs if it is an important, original, or non-standard component of the core methods in this research? Note that if the LLM is used only for writing, editing, or formatting purposes and does not impact the core methodology, scientific rigorousness, or originality of the research, declaration is not required.
    \item[] Answer: \answerYes{} 
    \item[] Justification: Yes, the paper describes the LLM backbone.
    \item[] Guidelines:
    \begin{itemize}
        \item The answer NA means that the core method development in this research does not involve LLMs as any important, original, or non-standard components.
        \item Please refer to our LLM policy (\url{https://neurips.cc/Conferences/2025/LLM}) for what should or should not be described.
    \end{itemize}

\end{enumerate}

\end{document}